\documentclass[11pt,a4paper]{article}
\usepackage{times,latexsym}
\usepackage{url}
\usepackage[T2A,T1]{fontenc}

    \usepackage[acceptedWithA]{tacl2021v1}
\newif\iftaclinstructions
\taclinstructionsfalse % AUTHORS: do NOT set this to true
\iftaclinstructions
\renewcommand{\confidential}{}
\renewcommand{\anonsubtext}{(No author info supplied here, for consistency with
TACL-submission anonymization requirements)}
\newcommand{\instr}
\fi

\iftaclpubformat % this "if" is set by the choice of options

\else

\fi

\usepackage{amsmath}
\usepackage{amssymb}

\usepackage[utf8]{inputenc}

\usepackage{graphicx}

\usepackage[russian,english]{babel}

\title{The Distribution of Phoneme Frequencies across the World's Languages:\\
Macroscopic and Microscopic Information-Theoretic Models}

\author{Fermín Moscoso del Prado Martín \\
  Department of Computer Science \\
  and Technology \\
  University of Cambridge, UK \\
  \texttt{fm611@cst.cam.ac.uk} \\
  \And
  Suchir Salhan \\
  Department of Computer Science\\
   and Technology \\
  University of Cambridge, UK \\
  \texttt{sas245@cst.cam.ac.uk} \\}

\begin{document}
\maketitle
\begin{abstract}
We demonstrate that the frequency distribution of phonemes across languages can be explained at both macroscopic and microscopic levels. Macroscopically, phoneme rank–frequency distributions closely follow the order statistics of a symmetric Dirichlet distribution whose single concentration parameter scales systematically with phonemic inventory size, revealing a robust compensation effect whereby larger inventories exhibit lower relative entropy. Microscopically, a Maximum Entropy model incorporating constraints from articulatory, phonotactic, and lexical structure accurately predicts language-specific phoneme probabilities. Together, these findings provide a unified information-theoretic account of phoneme frequency structure.
\end{abstract}

\section{Introduction}

The frequency distributions with which linguistic units occur are consequences of the cognitive mechanisms involved in language production. As such, understanding them can provide useful insights into the mechanisms by which language is processed and represented in the human mind. In this context, much research has investigated the frequency distributions of words across languages. This is a power-law distribution that is approximately constant across languages, first formalised by \citet{Condon:1928} (although it is often attributed to \citealp{Zipf:1932}). 

Despite the fundamental place of phonemes in language structure and processing, only a handful of studies have investigated the frequency distribution of phonemes \cite{Sigurd:1968,Good:1969,Martindale:etal:1996,Martindale:Tambovtsev:2007,MacklinCordes:Round:2020}. All these studies have considered the relationship between a phoneme's probability of occurrence to its \emph{rank}, that is, its ordinal position within the ordered list of phoneme probabilities for a given language. None of these approaches models why some specific phonemes are more frequent than others.

In what follows, we begin by providing a short overview of previous attempts at describing phoneme frequency distributions. We then describe three phoneme frequency datasets that we will use in our models. With these datasets, we develop a two-level account of phoneme frequency distributions. First, we show that at a macroscopic scale --in the sense of \citet{Mandelbrot:1957}-- the rank–frequency structure of phonemes in any language can be derived from a virtually parameter-free symmetric Dirichlet model, with its single concentration parameter systematically related to inventory size. We then turn to the microscopic level and demonstrate how specific phoneme probabilities can be accurately \emph{guessed} using the Principle of Maximum Entropy.

\section{Models of Phoneme Frequency Distributions}

Inspired by the power law distributions found for word types, all previous attempts at modelling the frequency distribution of phonemes --with the sole exception of \citet{Good:1969}-- have used some form of a power law. \citet{Sigurd:1968} explored directly applying Zipf's Law \cite{Condon:1928} to model the frequency $f(p)$ of a phoneme $p$ as a function of its \emph{rank} $r(p)$ and some constant parameter $k$, such that $f(p) \approx k/r(p)$. Sigurd found that this approach did not produce a good fit for the phoneme frequency distributions of five languages. To improve upon this fit, he suggested using a geometric series approximation of \citet{Mandelbrot:1957}'s corrected version of Zipf's Law (the  Zipf-Mandelbrot Law) with additional parameters, which resulted in improved fits to his data.

\citet{Martindale:etal:1996} found that the phoneme distribution was better fit using the Yule-Simon distribution \citep{Yule:1924,Simon:1955}. They show that the Yule-Simon model is in fact a generalisation of \citet{Sigurd:1968}'s geometric series sum. In a further study, \citet{Martindale:Tambovtsev:2007} found that the Yule-Simon model outperforms Sigurd's model in fitting the phoneme frequency distributions observed in 95 different languages.

Most recently, \citet{MacklinCordes:Round:2020} re-evaluated the adequacy of power laws for modelling phoneme frequency distributions. Using a sample of 166 Australian language varieties, they found that power law distributions --such as those used in the previous studies-- are unsuitable for modelling the right tail of the phoneme frequency distribution. Instead of a single distribution, they define a distribution piecewise, so that it corresponds to a power law on its left tail (i.e., the highest frequency phonemes), and a plain exponential tail for the right tail (i.e., the lowest frequency phonemes).

\begin{figure}[t]
  \includegraphics[width=\columnwidth]{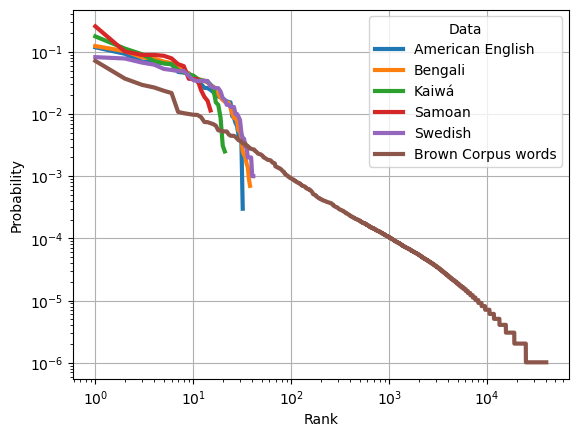}
  \caption{Rank-frequency plots for the five phoneme frequency distributions discussed in \citet{Sigurd:1968}, compared with the word frequency distribution in the Brown corpus. Note the logarithmic scales.}
  \label{fig:fig1}
\end{figure}

Power laws are typically appropriate for distributions spanning multiple orders of magnitude and often involving open-ended vocabularies \citep{Clauset:etal:2009}. Phoneme inventories are fundamentally different: they are small, closed sets, rarely exceeding a few dozen contrasts. As illustrated in Figure~\ref{fig:fig1}, phoneme frequency distributions do not exhibit the extended linear behaviour in double-logarithmic space characteristic of genuine power laws. Even proposals restricting power-law behaviour to part of the distribution therefore lack a principled justification. As is illustrated in the Figure, true power laws (e.g., word frequencies in the Brown corpus) form straight lines over several orders of magnitude in log–log space.

Surprisingly, even after correctly noticing that the right tails of phoneme frequency distributions do not correspond to power laws, \citet{MacklinCordes:Round:2020} nevertheless insist on fitting a power law to the \emph{left} tail of the distributions. This is difficult to justify, since power laws are typically appropriate only for open-ended right tails.

\section{Datasets}
\label{sec:data}

We will use three datasets of phoneme frequency distributions: The first dataset are the five phoneme frequency distributions (for American English, Bengali, Kaiw\'a, Samoan, and Swedish) discussed by \citet{Sigurd:1968}. These data were obtained by transcribing the tables provided in the original article. 

The second dataset is \citet{MacklinCordes:Round:2020}'s extensive set of phoneme frequency distributions for 166 Australian language varieties.\footnote{Downloaded from \url{https://zenodo.org/records/4104116} on May 1, 2025.} These are also high-quality inventories carefully curated by experts on each individual language. However, despite the large number of languages, and substantial genetic diversity (covering 19 different language families), Australian languages are known to span relatively little typological diversity in terms of phonemic inventories \citep{MacklinCordes:Round:2020}.

\begin{figure}[t]
  \includegraphics[width=\columnwidth]{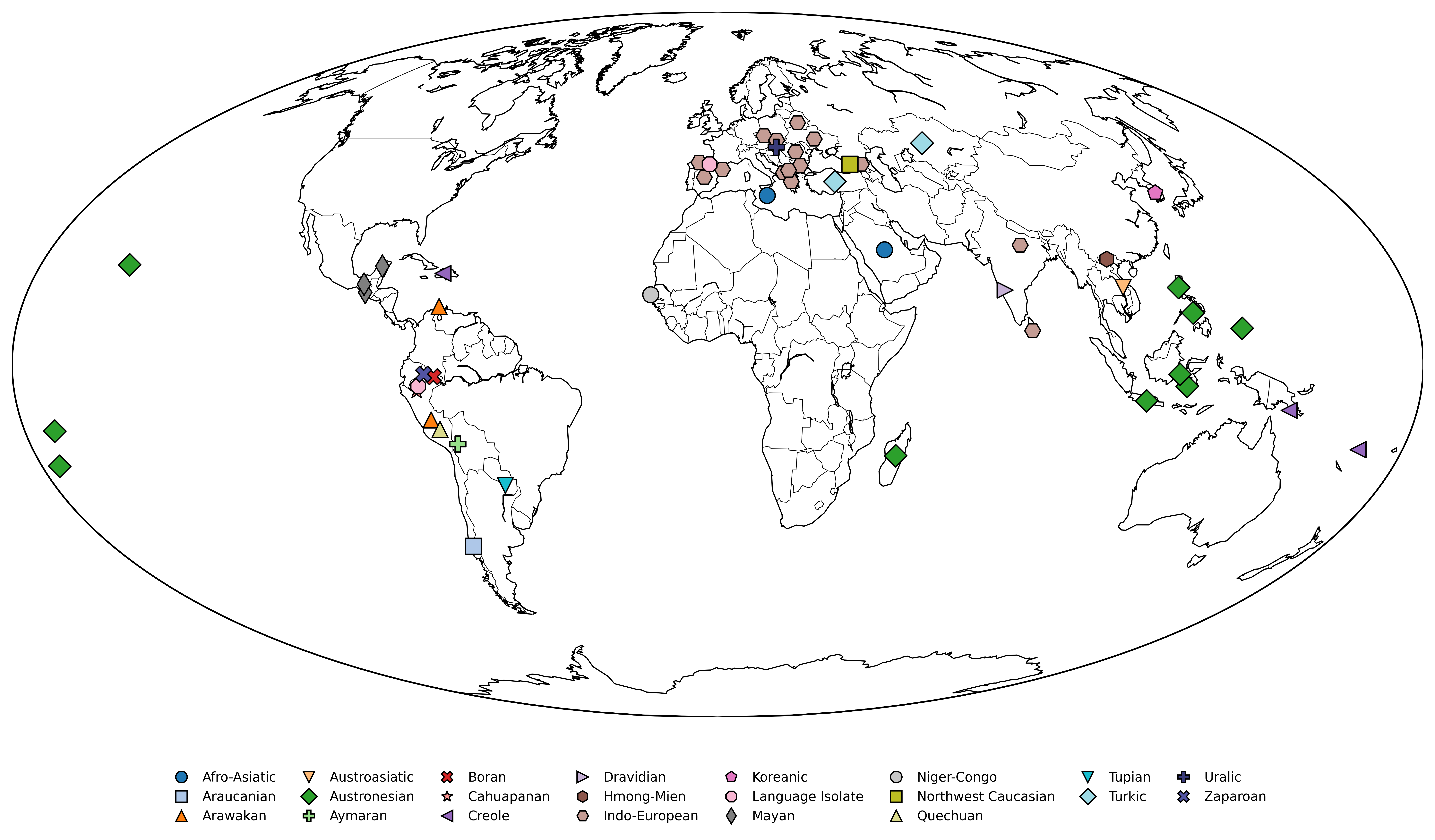}
  \caption{Geographic and genetic diversity of the languages included in the UDHR dataset.}
  \label{fig:map}
\end{figure}

Finally, in order to compensate for the limited typological and geographical diversity, we compiled our own dataset of phonemic frequency distributions. This was achieved semi-automatically by applying the automatic phoneme transcription available in the Cross-linguistic Phonological Frequencies Corpus (XPF; \citealp{XPF:2021})\footnote{We modified {XPF}'s algorithm so that the long and short versions of phonemes would always be counted as separate phonemes.} to different versions of the Universal Declaration of Human Rights ({UDHR}) distributed as a corpus in Python's Natural Language Toolkit \citep{NLTK:2009}. In this way, we obtained phoneme frequency distributions for a sample of 53 languages (see Appendix~\ref{ap:langs} for the list of languages), very widely distributed in genetic, geographic, and typological terms (see Figure~\ref{fig:map}). Evidently, what has been gained in typological and geographical diversity with respect to the Australian dataset, has been lost in terms of accuracy: The distributions  automatically computed from small written corpora cannot be compared to the carefully curated phonemic inventories. However, they provide broad genealogical coverage within a uniform domain and enable computation of the features required for the microscopic analyses below. As the macroscopic results below show, their phoneme distributions display the same structural patterns as the curated data, indicating that they capture the relevant statistical structure despite potential noise.

\section{The Macroscopic Distributions of Phonemes across Languages}

To find the possible distribution of phonemes in a given language, one can start from the trivial constraint that the probabilities of all its phonemes must add up to one. In other words, for any language with exactly $n$ distinct phonemes, all possible phoneme frequency distributions must lie within the $(n-1)$-simplex. Each possible distribution is just a multinomial. One can define a prior probability distribution over these multinomials. The natural choice for a prior is the conjugate distribution for a multinomial:  a Dirichlet distribution $\mathrm{Dir}(\boldsymbol{\alpha})$, where $\boldsymbol{\alpha} \in \mathbb{R}^n$ is a vector of $n$ parameters. Each point sampled from a Dirichlet distribution is a multinomial distribution of dimension $n$.

\begin{figure*}[th]
\begin{tabular}{ccc}
{\bf (a)} & {\bf (b)} & {\bf (c)} \\
  \includegraphics[width=0.31\linewidth]{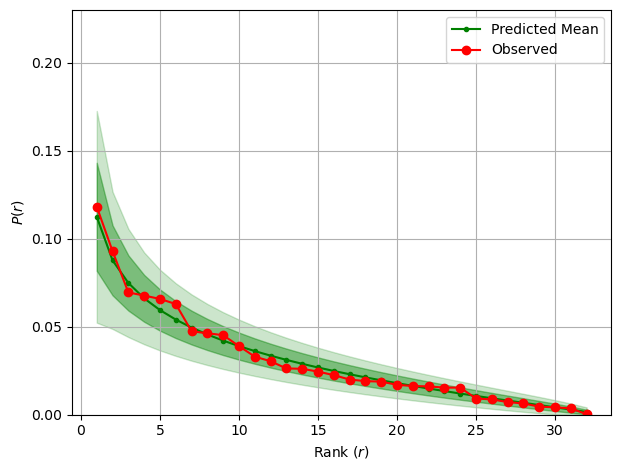} &
  \includegraphics[width=0.31\linewidth]{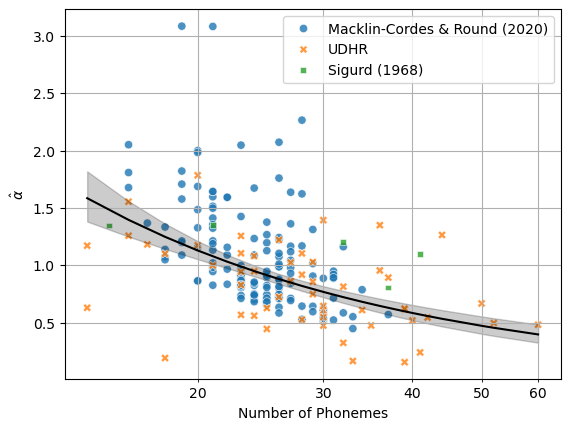} &
  \includegraphics[width=0.31\linewidth]{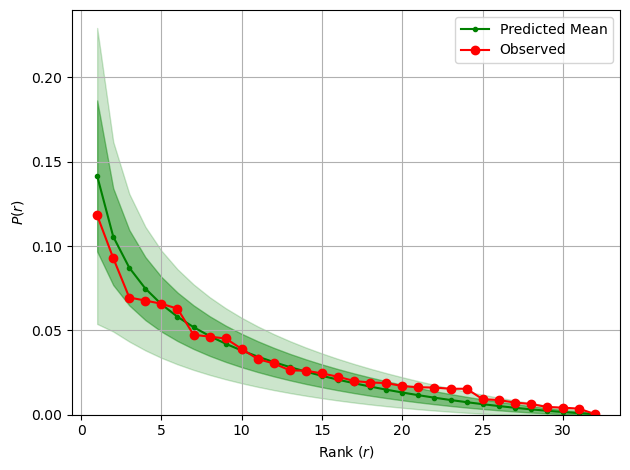} 
 \end{tabular}
  \caption {{\bf (a)} Rank-frequency plot for the American English phoneme frequency distribution (red dots and lines) discussed in \citet{Sigurd:1968}, overlaid with the predicted rank-frequency distribution for symmetric Dirichlet distributions with estimated concentration parameters $\hat\alpha$. The green lines are the predicted mean order statistics, the darker green shading denotes their standard deviations, and the lighter green shading denotes their 95\% C.I. {\bf (b)} Relationship between the size of the phonemic inventory (horizontal axis; note the logarithmic scale) with the estimated value of the concentration parameter $\hat\alpha$ (vertical axis) across the three datasets. Each point denotes the phonemic inventory for a language variety. The black line plots a doubly logarithmic linear regression (whose parameters are given in Equation~\ref{eq:prediction}) and the shaded area is its 95\% C.I. {\bf (c)} Rank-frequency plot for the American English  phoneme frequency distribution (red dots and lines) discussed in \citet{Sigurd:1968}, overlaid with its rank-frequency distribution reconstructed from its inventory size using Equation~\ref{eq:prediction}. The green lines are the predicted mean order statistics, the darker green shading denotes their standard deviations, and the lighter green shading denotes their 95\% C.I.}
  \label{fig:macroplots}
\end{figure*}

At the macroscopic scale, phoneme identity is abstracted away, so no phoneme-specific assumptions are introduced. Therefore, there is no reason why any phoneme should be a priori more or less probable than any other one. In terms of the Dirichlet distribution this means that the $\alpha_i$ components of the vector $\boldsymbol{\alpha}$ should all be equal, $\alpha_1 = \alpha_2 = \ldots = \alpha_n = \alpha > 0$. In this situation, we have a single parameter distribution, a \emph{symmetric Dirichlet distribution} with concentration parameter $\alpha$. Values of $\alpha=1$ indicate full ignorance about the distribution of phonemes, with all multinomial distributions being equally probable (i.e., it is a uniform distribution on the $(n-1)$-simplex). On the one hand, values of $0< \alpha < 1$ would indicate that  skewed distributions --when a few phonemes are much more frequent than the rest-- are most likely. On the other hand, $\alpha>1$ would make the more even phoneme distributions more probable. We can use the marginal distribution to compute the expected values of its rank positions using the \emph{order statistics} \citep{David:Nagaraja:2003} of the distribution: The first order statistic in a sample of size $n$ corresponds to the phoneme ranked n, the second order statistic to rank $n-1$, and so on, with the $n$-th order statistic being the most frequent phoneme (rank 1). 

\citet{Good:1969} observed that the frequencies of English phonemes matched remarkably well the order statistics corresponding a symmetric Dirichlet distribution with concentration parameter $\alpha = 1$,\footnote{Good describes it using the `stick breaking' process where the breaks are determined uniformly, which is equivalent to the symmetric Dirichlet with $\alpha=1$.}   and the author states that this holds for other languages as well. Good could not provide any explanation for this empirical observation. Incidentally, although not explicitly calling it a Dirichlet distribution, this $\alpha=1$ case is also what \citet{GuseinZade:1988} found to be a good model of Russian letter frequencies. On the lack of any additional information, one should expect the Laplace non-informative prior ($\alpha=1$) to be a rather good guess (i.e., it is a maximum entropy prior on the space of distributions). However, \citet{Martindale:etal:1996} found that the $\alpha=1$ case was not the best model for phoneme frequencies.

\subsection{Concentration parameter values}

For each of the phoneme frequency distributions described in Section~\ref{sec:data}, we estimated the optimal value of the concentration parameter $\hat\alpha$ and associated order statistics (using the robust methods described in Appendix~\ref{sec:fitmethod}). Figure~\ref{fig:macroplots}a illustrates the accuracy of the fits obtained; it overlays the rank-frequency plot of the phoneme frequencies for the American English dataset reported in \citet{Sigurd:1968}, with what would be expected from a symmetric Dirichlet distribution with the corresponding estimated value of the concentration parameter ($\hat\alpha$). More generally, the quality of the fits was remarkably good for all the phoneme frequency distributions in all three datasets.

As shown in Figure~\ref{fig:macroplots}b, there is substantial cross-language variability in the optimal value of the concentration parameter, with some languages tending to more uniform distributions (with $\alpha > 1$) and others tending to more skewed ones (with $\alpha<1$). However, a striking regularity stands out: Across datasets, there is a clear negative correlation ($\beta =  -0.95 \pm 0.11,\,t=-8.46,\,p<.0001$ on a log-log linear regression) between the size of a language's phonemic inventory and its corresponding concentration parameter estimate, i.e., languages having more phonemic contrasts also exhibit lower values of $\hat\alpha$. The reliable relationship between phonemic inventory size and concentration parameter value enables us to \emph{predict} the value of $\alpha$ for any language solely on the size of its phonemic inventory. Fitting a linear regression predicting the logarithm of $\alpha$ as a function of the logarithm of $n$,\footnote{The prediction needs to be made on double logarithmic scale because both $\alpha$ and $n$ are positive by definition.} one can estimate $\alpha$ for a language with $n$ contrasts as the regression curve depicted in Figure~\ref{fig:macroplots}b
\begin{equation}
\hat{\alpha}(n) \approx 19.47 \cdot n^{-.95} \label{eq:prediction}
\end{equation}
Crucially, even when the regression includes dataset origin and its interaction with inventory size, the negative relationship between concentration parameter and inventory size remains strong and highly significant. This indicates that the scaling relation is not an artefact of dataset composition, but a robust structural property of phoneme frequency distributions. It predicts a range of $\alpha$ values going from a minimum of $.16$ for East Taa (160 phonemes; cf., \citealp{Green:Moran:2019}) to a maximum of $2.00$ for Central Rotokas (11 contrastive phonemes; cf., \citealp{Robinson:2006}). Once one has predicted the value of $\alpha$, one can directly reconstruct the predicted rank-frequency plot using the order statistics (see Appendix~\ref{sec:order}). Figure~\ref{fig:macroplots}c illustrates the accuracy of reconstructing a phoneme frequency distribution using Equation~\ref{eq:prediction}. Similar performance was obtained in the frequency distributions of all three datasets.

\subsection{The `Compensation Hypothesis'}

The symmetric Dirichlet distribution is a distribution over distributions. An $n$-dimensional multinomial distribution $X$ sampled from a symmetric Dirichlet distribution with concentration $\alpha$ is expected to have an entropy \cite{Shannon:1948},
\begin{equation}
\mathbb{E}_{n,\alpha}[H(X)] = \psi(\alpha n + 1) - \psi(\alpha +1) \label{eq:entropy1}
\end{equation}
where $\psi$ denotes the digamma function. When $\alpha \to \infty$ the expected entropy converges to the maximum possible entropy for $n$ options, $H_\text{max}[n] = \log n$. That $\hat\alpha$ is negatively correlated with the inventory size $n$ entails that the \emph{relative} entropy (i.e., the proportion of $H(X)$ relative to $H_\text{max}[n]$) of phoneme frequency distributions decreases as the size of the phonemic inventory grows. Using Equation~\ref{eq:prediction}, relative entropies are expected to range from 71\% for East Taa, to a quite uniform-like 91\% for Central Rotokas.

The entropy of the phoneme distribution reflects the average informational cost of processing phonemes \citep{Cherry:etal:1953}, and inventory size has traditionally served as a proxy for phonological complexity \citep{Trubetzkoy:1939,Hockett:1955}. However, we observe a trade-off: as a language’s phonemic inventory increases, the relative entropy of its phoneme distribution decreases, so that the increase in the information processing cost of phonemes is attenuated. This pattern aligns with the `Compensation Hypothesis' \citep{Hockett:1955,Martinet:1955}: increased complexity in one domain of the language system is offset elsewhere. Many have found evidence for this hypothesis involving either the properties of phoneme sequences, or in the microscopic properties of the phonemes (e.g., \citealp{Moran:Blasi:2014,Pimentel:etal:2020}). Remarkably, this compensation is already visible in unigram phoneme distributions.

\section{The Microscopic Distribution of Phonemes across Languages}

The \emph{Entropy Concentration Theorem} \citep{Jaynes:1989}  states that the probability of probability distributions for a given phenomenon peaks strongly around those distributions with the maximum \emph{possible} entropy; by possible it is meant that they \emph{satisfy all constraints} that govern the phenomenon. Distributions even slightly less entropic than the maximum are highly improbable because --informally speaking-- there are much fewer possible low entropy distributions than there are higher entropy ones. Jaynes discusses that, whenever one finds that an observed distribution's entropy is below its maximum, this implies that one is missing some constraints. When all relevant constraints have been identified, one finds that the entropy is indeed within a small range of its maximum, and the problem has then been truly understood.

The macroscopic analyses in the previous section revealed that, even for the languages with the smallest possible phonemic inventories, the entropy of their phoneme distributions remains below its maximum value for a given phonemic inventory size. Following \citet{Jaynes:1989}, such relatively low entropies indicate that there are some additional factors constraining the distributions of phonemes. These constraints can be understood at the `microscopic' level, in the sense that they relate to the reasons that determine why specific phonemes are more or less common within a given language, e.g., why is phoneme /n/ more frequent than /d/ in English? If we manage to spell out these additional constraints in sufficient detail, we could use the method of Maximum Entropy \citep{Jaynes:1957a,Jaynes:1957b} for \emph{guessing} the distribution of specific phonemes in a given language with a high degree of accuracy.

\subsection{The Principle of Maximum Entropy}

Let $\text{p}(p)$ be a distribution over possible phonemes $p$ within a given language, and let $\{f_k(p)\}$ be given feature functions of each phoneme, whose expectations across phonemes in a language are given by $c_k$. The maximum--entropy distribution is defined by the variational problem
\begin{equation}
\text{p}^\star =
\begin{aligned}[t]
\arg\max_{\text{p}(p)} \;& - \sum_p \text{p}(p)\log \text{p}(p)\\
\text{s.t.}\;& \sum_p \text{p}(p)=1,\\
& \sum_p \text{p}(p)f_k(p)=c_k\;\forall k 
\end{aligned} \label{eq:maxent}
\end{equation}
The first constraint is plain normalisation, and the remaining constraints fix the expectations of the feature functions. Introducing Lagrange multipliers $\lambda_k$, the solution takes the classical Gibbs-Boltzman form. In logarithmic scale we can write this solution as,
\begin{equation}
\log \text{p}^\star(p) = \lambda_0  + \sum_k \lambda_k f_k(p) \label{eq:maxentsol}
\end{equation}
where the $\lambda_k$ are chosen so that the constraints are satisfied and the entropy maximised. Notice that the Lagrange multiplier values can be interpreted as coefficients in a linear regression predicting the log phoneme probabilities. Unlike regression, however, rather than estimating coefficients from observed probabilities, Maximum Entropy derives the least-structured distribution consistent with the constraints. The resulting Lagrange multipliers are therefore not \emph{fitted} to phoneme probabilities, but inferred from constraint expectations, making the approach generative rather than descriptive.

\subsection{Constraints on phoneme frequencies}

Our task in maximum entropy modelling involves identifying the relevant constraints that provide the $f_k$ and $c_k$ terms in Equations~\ref{eq:maxent} and \ref{eq:maxentsol}. Rather than providing an extensive list of the complete set of constraints that can affect phoneme frequency distributions, the goal of this study is demonstrating how these constraints arise from multiple levels, ranging from language-independent physical factors to higher level linguistic ones. We will illustrate the diversity of constraints at play using three concrete aspects.

\subsubsection{Physical factors}
\label{sec:physical}

Different phonemes involve different physical `costs'. On the one hand, the articulatory movements involved in producing different phonemes are likely to differ in terms of their energy costs. On the other hand, in language comprehension, some phonemic contrasts might be physically easier to perceive than others. Rather than attempting to model these costs explicitly (e.g., \citealp{Maddieson:1984,Ladefoged:Maddieson:1996}), we propose measuring them by proxy. Across languages, phonemic contrasts involving higher energy costs should be expected to be less common than those requiring less energy (e.g., \citealp{Boersma:1998}). This results in contrasts that are more common across languages being also more frequent within the individual languages in which they occur \citep{Stemberger:Bernhardt:1999,Gordon:2016}.

The above entails that we can operationalise the physical cost of specific phonemic contrasts by the number of distinct languages in which each occurs, their \emph{incidence frequencies} \citep{Gotelli:Chao:2013} across the world's languages. For each phoneme in the phonemic inventories in our UDHR dataset, we computed its cross-linguistic incidence probability according to PHOIBLE 2.0.\footnote{Slight differences in notation between XPF and PHOIBLE 2.0 made it impossible to compute incidence frequencies for a small minority of phonemes. Nevertheless, we succeeded in computing the incidence frequencies for a median of 99\% of each language's phonemes, and at least 89\% of the phonemes in every language. Unmatched phonemes were excluded from further analysis.} For each phoneme $p$, we used this incidence probability $\text{p}_i(p)$ to compute a proxy for the phoneme's physical cost,
\begin{equation}
\text{cost}(p) = -\log \text{p}_i(p) \label{eq:physcost}
\end{equation}
indicating that phonemes that are rarer across languages have higher physical costs.

\subsubsection{Phonotactic factors}
\label{sec:phonotactic}

Phonemes are typically embedded within linguistic sequences, which are known to include a substantial degree of redundancy \citep{Shannon:1951}. Phonemes that are more predictable from their contexts are more likely to be elided \citep{CohenPriva:2015}. Diachronically, these elisions can lead to the outright dropping of the phoneme in those contexts where it is very predictable (e.g., \citealp{Martinet:1952,Ohala:1993}). This could entail a rather counterintuitive effect: One would expect more frequent phonemes to be more predictable; in terms of phoneme $n$-grams, predictable phonemes occur in relatively high frequency $n$-grams. The unigram frequencies of those phonemes are just the sums of the frequencies of the $n$-grams in which they are found. We would then expect their unigram frequency to be also relatively high. However, the diachronic considerations above suggest that in fact the more predictable a specific phoneme is within a language, the less frequent it becomes.

In order to measure the degree to which a phoneme occurs in predictable contexts, we used the \emph{segmental information} measure \citep{VanSon:Pols:2002,VanSon:Pols:2003a,VanSon:Pols:2003b}. For a given phoneme $p$ preceded (within an individual word) by the sequence of phonemes in a given word onset $o$, the segmental information is given as the logarithmic ratio of the frequency of that specific phoneme in that context, to that of any phoneme $P$ in that same context,
\begin{equation}
I_s(p;o) = \log \frac{\text{Frequency}(o+P)}{\text{Frequency}(o+p)} \label{eq:seginfo}
\end{equation}
This measures how surprising it is to find a given phoneme within a specific context. If the phoneme is the only possible continuation in that context, its segmental information in that context is zero, and it becomes higher as the relative probability of that continuation decreases. One can average the sequential information across all word onsets $\mathcal{P}^*$ using a phoneme inventory $\mathcal{P}$,
\begin{align}
I_s(p) & = \left< I_s(p;o) \right>_{\mathcal{P}^*} \nonumber \\
& = \sum_{o \in \mathcal{P}^*} \text{p}(o|p) I_s(p;o) \label{eq:seginfo2}
\end{align}
to obtain a measure of the phoneme's informativity. We computed this measure for each phoneme in each of the languages in our UDHR dataset, averaging over the contexts within the corresponding versions of the UDHR.

\subsubsection{Higher level linguistic factors}
\label{sec:word}

One can also ask whether `higher' tiers of language organisation are reflected in the distribution of phonemes. Ultimately, the main function of phonemes --what makes them contrastive-- is to distinguish between different words. The lexical organisation level is therefore a good candidate for influencing the phoneme frequency distributions. The uncertainty of the distribution of words can be measured by its entropy, the \emph{lexical diversity} \citep{Spokoyny:etal:2016,Moscoso:2017}.  As shown by psycholinguists \citep{WMW:1987}, each of a word's phonemes sequentially contributes to reducing this uncertainty. The amount of lexical uncertainty reduction by a specific phoneme $p$ occurring after an onset context $o$ is what we can term the phoneme's \emph{lexical information gain} in prefix context $o$,
\begin{equation}
I_\ell(p;o) = H(W|o) - H(W|o+p) \label{eq:lexinfo}
\end{equation}
such that the sum of all lexical information gains adds up to the lexical diversity. Each phoneme has an individual contribution to the overall lexical diversity, which is weighted by the phoneme's probability,
\begin{align}
H(W) & = \sum_{p \in \mathcal{P}} \sum_{o \in \mathcal{P}^*} \text{p}(o,p) \, I_\ell(p;o) \nonumber \\
&  = \sum_{p \in \mathcal{P}} \text{p}(p) \sum_{o \in \mathcal{P}^*} \text{p}(o|p) \, I_\ell(p;o) \nonumber \\
&  = \sum_{p \in \mathcal{P}} \text{p}(p) \, I_\ell(p) \label{eq:lexinfo2}
\end{align}
where $I_\ell(p)$ denotes a phoneme's  information gain about word identity averaged across prefix contexts. We can also define the entropy of words conditional on phonemes as the average uncertainty about words containing a phoneme $p$,
\begin{equation}
H(W|\mathcal{P}) = \sum_{p \in \mathcal{P}} \text{p}(p)\, H(W|p) \label{eq:lexinfo4}
\end{equation}
We refer to $H(W|p)$ as the \emph{lexical diversity conditioned on phoneme $p$}. Crucially, the lexical diversity can also be decomposed as
\begin{equation}
H(W) = I(W;\mathcal{P}) + H(W|\mathcal{P}) \label{eq:lexinfo5}
\end{equation}
where $I(W;\mathcal{P})$ denotes the mutual information between words and phonemes. From the above, an unbiased estimator of a phoneme’s lexical information gain is the lexical diversity conditioned on that phoneme plus a non-negative constant
\begin{align}
 I(W;\mathcal{P}) & = \sum_{p \in \mathcal{P}} \text{p}(p) \, \big( I_\ell(p) - H(W|p) \big) \geq 0 \nonumber \\
I_\ell(p) & \approx I(W;\mathcal{P}) + H(W|p) \label{eq:lexinfo6}
\end{align}
This estimator is not, in general, consistent (i.e., it does not converge on the true value). However, the mutual information term is expected to be much smaller than the conditioned diversity (i.e., one should expect $I(W;\mathcal{P}) \leq  H(\mathcal{P}) \ll H(W|p)$ because in any human language the number of possible words far exceeds the number of distinct phonemes). Hence, we can use $H(W|p)$ to capture most of the variability in $I_\ell(p)$. We therefore estimated $H(W|p)$ for each phoneme in each language of the UDHR dataset using \citet{Chao:etal:2013}'s smoothing method.

\subsection{Maximum entropy results}

For each phoneme $p$ in each of the languages $L$ in the UDHR dataset, we have three feature functions, $f_1(p,L)$, $f_2(p,L)$, and $f_3(p,L)$ corresponding to the factors described in sections~\ref{sec:physical}, \ref{sec:phonotactic}, and \ref{sec:word}, respectively. We can use these feature functions to compute, for each language, the expected values of the features,
\begin{equation}
c_i(L) = \sum_{p \in L} \text{p}_L(p) f_i(p,L),\; \text{for } i=1,2,3 \label{eq:expectations}
\end{equation}
where $\text{p}_L(p)$ denotes the probability with which phoneme $p$ occurs in language $L$. These values allow us to set for each language $L$ the corresponding version of the maximum entropy variational problem (Equation~\ref{eq:maxent}) and its solution (Equation \ref{eq:maxentsol}). The values of the resulting Lagrange multipliers $\lambda_1(L)$, $\lambda_2(L)$ and $\lambda_3(L)$ for each language are then obtained by numerical maximisation of the equation. 

\begin{figure*}[t]
\begin{tabular}{ccc}
{\bf (a)} & {\bf (b)} \\
  \includegraphics[width=.48\linewidth]{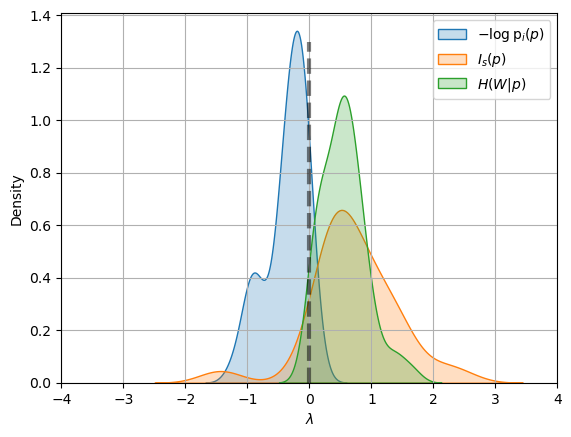} &
  \includegraphics[width=.48\linewidth]{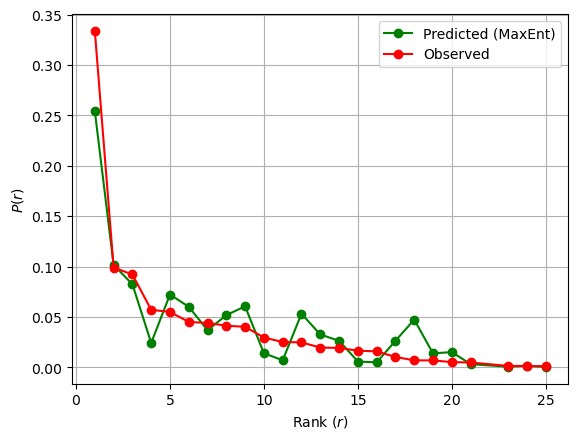} \\
{\bf (c)} & {\bf (d)}  \\
  \includegraphics[width=.48\linewidth]{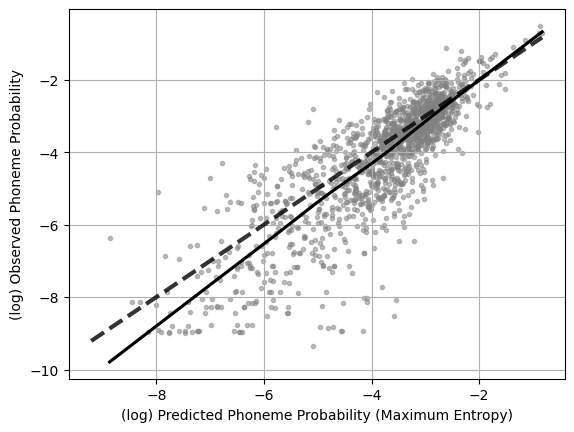} &
  \includegraphics[width=.48\linewidth]{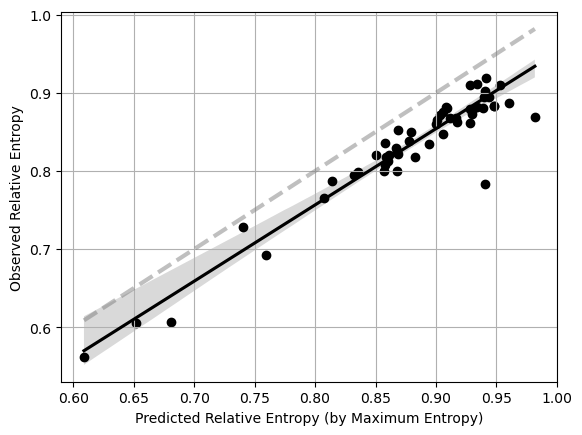} 
\end{tabular}
  \caption {{\bf (a)}  Distributions (kernel density estimates) of the estimated values of the maximum entropy Lagrange multipliers across the UDHR datasets. {\bf (b)} Comparison between the observed (red) and maximum entropy guessed (green) probabilities of the phonemes in the Abkhaz UDHR dataset (the horizontal axis plots the ranks of the observed data). {\bf (c)} Comparison between the maximum entropy guessed (horizontal axis) and observed (vertical axis) probabilities of the phonemes across all UDHR datasets. The solid line plots a non-linear regression (lowess) and the dashed line plots the identity (note logarithmic scales). {\bf (d)} Comparison between the maximum entropy guessed (horizontal axis) and observed (vertical axis) relative entropies of the phoneme distributions across all UDHR datasets. The solid line plots a linear regression (the shading denotes its 95\% C.I.) and the dashed line plots the identity.}
   \label{fig:maxentres}
\end{figure*}

\subsubsection{Lagrange multiplier values}

The  distributions of the estimated Lagrange multiplier values are plotted in Figure~\ref{fig:maxentres}a. As expected, maximum entropy predicts a negative effect (i.e., $\lambda<0$ for 94\% of languages) of the physical cost of phonemes on their probability within a language. In other words, one should expect phonemes that are rarer across languages to be rare within a language as well. On the other end of the plot, the phonotactic surprisal has a positive effect ($\lambda > 0$ for 94\% of languages). This confirms our counterintuitive prediction that phonemes that occur in more informative or surprising sequences tend to actually be more frequent, possibly as a trace of a diachronic effect by which predictable phonemes are dropped. Finally, the Lagrange multiplier for the lexical information gain proxy is positive for 98\% of languages in the UDHR dataset. In other words, phonemes that contribute more information towards lexical identification are also expected to be more frequent. Higher level linguistic factors such as those related to lexical aspects are directly reflected in the unigram phoneme frequency distribution.  In addition, there was a slight correlation (Pearson's $r=.36,\,t(51)=2.77,\,p=.0078$)  between the (mostly negative) multipliers corresponding to the physical cost and those (mostly positive) of the lexical aspects. This indicates that, in those languages whose phoneme probabilities are most sensitive to lexical discriminability, the probabilities are expected to be less sensitive to the actual physical cost of the phonemes.

%\subsubsection{Regression analyses}  We fit a linear mixed effect model \citep{Bates:etal:2015} predicting the log observed phoneme probabilities. We found significant fixed effects of the physical cost ($\beta = -.25 \pm .03,\, t=-8.68,\, p<.0001$), the phonotactic surprisal ($\beta = 1.33 \pm .24,\, t=5.55,\, p<.0001$), and the lexical information gain proxy ($\beta = .79 \pm .03,\, t=25.70,\,p<.0001$), as well as an interaction between the physical cost and the phonotactic surprisal that worked as a slight attenuation of these effects ($\beta=-.065\pm.025,\,t=-2.61,\,p=.007$). We found significant random slopes of the physical cost and the lexical information gain by language identity, indicating cross-language variability in the degree of sensitivity to these two factors (that are nevertheless present overall). To avoid multicollinearity, the log phonemic inventory size was residualised out of each of the three fixed effect predictors prior to fitting the regression. 

\subsubsection{Predicted phoneme probabilities} 

The phoneme probabilities guessed by maximum entropy matched rather accurately the phoneme probabilities actually observed. Figure~\ref{fig:maxentres}b compares the true and guessed frequencies of Abkhaz phonemes. In general, across languages, there is substantial coincidence between the guessed and true probabilities. This is highlighted by Figure~\ref{fig:maxentres}c, which compares guessed and true phoneme probabilities across all languages in the UDHR dataset. The correlation between the true and guessed probabilities is so clear that the non-linear regression (solid line) follows closely the identity (dashed line).

\subsubsection{Compensation revisited} 

We can now return to the macroscopic observation that the larger the phoneme inventories, the lower the relative entropy of their distribution. Figure~\ref{fig:maxentres}d shows that the relative entropies of the guessed distributions very closely match their observed counterparts. The relationship between inventory size and relative entropy must therefore have been captured. The figure also shows that the entropies of the observed distributions are generally lower --but only slightly so-- than their guessed counterparts (i.e., they are below the identity line). Following \citet{Jaynes:1989} this signals that some additional constraints are required to fully explain the distributions, but we are already remarkably close.

The influence of the predictive factors does not depend on the phonemic inventory size --the Lagrange multiplier values and phonemic inventory sizes are all uncorrelated. However, the average values of these factors (the $c_k$ in Equation~\ref{eq:maxent}) are indeed related to the inventory size. Some of these correlations are what one would expect by mere chance: The more phonemes one samples, the more likely one is to encounter improbable ones. Consequently, there is a positive correlation between the (log) phonemic inventory size and the average value of the physical cost (Pearson's $r=.36,\,t(51)=2.78,\,p=.0075$). Other correlations are, however, less evident: There is a negative correlation between the (log) phonemic inventory size and the average phonotactic surprisal of phonemes (Pearson's $r=-.36,\,t(51)=-2.80,\,p=.0072$). This indeed suggests a non-trivial compensation between phonemic inventory size and phonotactic structure, i.e., languages with larger phoneme inventories tend to have less unpredictable phonotactic structure. This is consistent with previous findings of compensation in phonotactic structure \citep{Moran:Blasi:2014,Pimentel:etal:2020,Pimentel:etal:2021}.

\section{Conclusions}

This study shows that phoneme frequency distributions can be explained at two complementary levels, one macroscopic and one microscopic. At the macroscopic level, rank–frequency patterns closely follow the order statistics of a symmetric Dirichlet distribution whose single concentration parameter scales predictably with inventory size. This makes the distribution virtually parameter-free. Given a known phonemic inventory size, one can reconstruct the rank-frequency distribution with substantial accuracy using just Equation~\ref{eq:prediction}. This provides a principled alternative to previous attempts at modelling phoneme frequencies \citep{Sigurd:1968,Good:1969,Martindale:etal:1996,Martindale:Tambovtsev:2007,MacklinCordes:Round:2020}.

The negative relationship between inventory size and the concentration parameter entails that languages with larger phoneme inventories exhibit lower relative entropy in their frequency distributions. This finding offers quantitative support for the Compensation Hypothesis \citep{Hockett:1955,Martinet:1955}: Increases in the number of contrasts are compensated with decreases in distributional evenness. In contrast with previous studies (e.g., \citealp{Moran:Blasi:2014,Pimentel:etal:2020,Pimentel:etal:2021}), this compensation can be observed directly from just the unigram phoneme distributions.

At the microscopic level, deviations from maximum entropy reveal the presence of meaningful constraints \citep{Jaynes:1989}. This framework introduces a novel methodology for analysing linguistic phenomena by identifying the constraints that shape probability distributions. To our knowledge, this is the first account of phoneme frequency distributions across languages at such a detailed level. The results show that unigram phoneme frequencies simultaneously reflect physical pressures, phonotactic structure, and influences from higher tiers of organisation such as the lexicon.

%\section*{Acknowledgments}

%We are indebted to Paul Siewert for comments on a previous version of this manuscript.

%\nocite{Maslova:2000,Cysouw:2010}%Maddieson:2006,Maddieson:2007}

% Custom bibliography entries only

\bibliography{custom}
\bibliographystyle{acl_natbib}

\appendix

\section{Robust Estimation of the Symmetric Dirichlet Distribution}
\label{sec:fitmethod}

\subsection{Smoothed entropy estimator of $\hat{\alpha}$}

Although it is possible to fit the concentration parameter $\alpha$ for a symmetric Dirichlet distribution directly using traditional maximum likelihood or Bayesian methods, we chose to use an indirect method. This allowed us to explicitly correct the sampling error in our distribution. As indicated in Equation~\ref{eq:entropy1}, the value of $\alpha$ is directly related to the expected entropy \citep{Shannon:1948} of a distribution sampled from it. We can take advantage of this relationship, estimating $\alpha$ from the entropy of the observed phonemic frequency distributions. These distributions are, however, samples from their true values. As such, they are expected to be underestimates of their true entropy values \citep{Miller:1955}. This provides an opportunity for correcting the sampling effect without having to adjust the probabilities themselves. Given a sample from a distribution --even under severe undersampling-- the Chao-Wang-Jost estimator (CWJ; \citealp{Chao:etal:2013}) provides an accurate and unbiased estimate of the true entropy of the underlying distribution. We can therefore compute the CWJ estimate from the observed frequency distribution, which we will denote by $\hat{H}$. If we know that there are $n$ distinct phonemes in a given language, the estimated value of $\alpha$ (which we denote by $\hat\alpha$) is the solution to the non-linear equation
\begin{equation}
\hat{H} = \psi(n\hat\alpha + 1) - \psi(\hat\alpha + 1)
\end{equation}
where $\psi$ denotes the digamma function. Although this equation does not have a closed-form analytical solution, it is quite easily solved numerically.

\subsection{Estimation of order statistics for the symmetric Dirichlet distribution}
\label{sec:order}

For an $n$-dimensional symmetric Dirichlet distribution with concentration parameter $\alpha$, its marginals follow a  $\text{Beta}(\alpha,(n-1)\alpha)$ distribution. This has a probability density function (PDF),
\begin{equation}
f(x) = \frac{x^{\alpha-1} (1-x)^{(n-1)\alpha-1}}{\text{B}\left(\alpha,(n-1)\alpha\right)} \label{eq:betapdf}
\end{equation}
where $\text{B}$ denotes the beta function. The corresponding cumulative density function (CDF) given in terms of the regularised incomplete beta function
\begin{equation}
F(x) = \frac{\int_0^x t^{\alpha-1} (1-t)^{(n-1)\alpha-1}\, \text{d}t}{\text{B}\left(\alpha,(n-1)\alpha\right)} \label{eq:betacdf}
\end{equation}

Given a PDF and CDF, the $r$-th out of $n$ order statistic is distributed with a known PDF \citep{David:Nagaraja:2003},
\begin{multline}
f_{(r,n)}(x) = \frac{n!}{(r-1)!(n-r)!} f(x) \\ [F(x)]^{r-1}[1-F(x)]^{n-r} \label{eq:orders}
\end{multline}
where one can plug in the Equations~\ref{eq:betapdf} and \ref{eq:betacdf}. One can use this distribution for estimating the expected means of order statistics,
\begin{equation}
\mu_{(r,n)} = \mathbb{E}_{f_{(r,n)}(x)}[x] = \int_0^1 x \, f_{(r,n)}(x)\, \text{d}x \label{eq:mu}
\end{equation}
and their variances,
\begin{multline}
\sigma^2_{(r,n)} = \text{Var}_{f_{(r,n)}(x)}[x] = \\ \int_0^1 \left(x-\mu_{(r,n)}\right)^2 \, f_{(r,n)}(x)\, \text{d}x \label{eq:sigma}
\end{multline}
The integrals in Equations~\ref{eq:mu} and \ref{eq:sigma} are not easily solvable for general values of $\alpha$,\footnote{For the special case of $\alpha=1$, a closed-form solution in terms of harmonic numbers is given by \citet{Whitworth:1901}, which was used by \citet{Good:1969} and independently approximated by \citet{GuseinZade:1988}.} but they converge rapidly using simple numerical integration methods.

\section{Languages in the UDHR dataset}
\label{ap:langs}

\begin{table}[h]
\centering
\small
\setlength{\tabcolsep}{8pt}
\renewcommand{\arraystretch}{1.05}
\begin{tabular}{lll}
Abkhaz & Haitian Creole & Ossetian \\
Albanian & Hawaiian & Palauan \\
Arabela & Hiligaynon & Quechua (Cuzco) \\
Arabic & Hindi & Romanian \\
Asturian & Hmong & Samoan \\
Aymara & Hungarian & Sinhala \\
Basque & Iloko & Slovak \\
Belarusian & Indonesian & Spanish \\
Bislama & Javanese & Tok Pisin \\
Bora & Kannada & Tongan \\
Buginese & Kazakh & Turkish \\
Bulgarian & Korean & Tzotzil \\
Candoshi & Macedonian & Ukrainian \\
Catalan & Malagasy & Vietnamese \\
Chayahuita & Maltese & Wayuu \\
Czech & Mam & Wolof \\
Greek & Mapudungun & Yucatec Maya \\
Guarani & Nomatsiguenga &  \\
\end{tabular}
\end{table}

\end{document}